%% file: main.tex
\documentclass{article}

\usepackage[dblblindworkshop, final]{neurips_2025}

\usepackage[utf8]{inputenc} 
\usepackage[T1]{fontenc}    
\usepackage{hyperref}       
\usepackage{url}            
\usepackage{booktabs}       
\usepackage{amsfonts}       
\usepackage{nicefrac}       
\usepackage{microtype}      
\usepackage{xcolor}         

\usepackage{booktabs}						
\usepackage{multirow}						
\usepackage{amsfonts}						
\usepackage{graphicx}						
\usepackage{duckuments}						

\usepackage{algorithm}
\usepackage{algorithmic}

\usepackage[utf8]{inputenc} 
\usepackage[T1]{fontenc}    
\usepackage{url}            
\usepackage{booktabs}       
\usepackage{amsfonts}       
\usepackage{nicefrac}       
\usepackage{microtype}      
\usepackage[dvipsnames]{xcolor}         
\usepackage{graphicx}
\usepackage{multirow}
\usepackage{color}
\usepackage{amsmath}
\usepackage{pifont}
\usepackage{subcaption} 
\usepackage{xspace}
\usepackage{bm}

\usepackage{amsmath}
\usepackage{amssymb}
\usepackage{mathtools}
\usepackage{amsthm}
\usepackage{xspace}
\usepackage{multirow}
\usepackage{colortbl}

\usepackage[capitalize,noabbrev]{cleveref}

\theoremstyle{plain}
\newtheorem{theorem}{Theorem}

\theoremstyle{definition}
\newtheorem{definition}{Definition}

\theoremstyle{remark}

\title{Nonlinear Laplacians Improve Signed-Directed Graph Learning}

\input{Styles/chapters/000authors}

\begin{document}

\maketitle

\input{chapters/00_abstract}

\input{chapters/01_Introduction}

\input{chapters/03_Spectheory}

\input{chapters/04_orgGNN}

\input{chapters/05_Experiments}

\input{chapters/06_Conclusion}
\bibliographystyle{unsrtnat}
\bibliography{reference}
\newpage 
\appendix
\input{chapters/02_RelatedWork}

\input{chapters/x1_appendix1}

\newpage

\end{document}

%% file: Styles/chapters/000authors.tex
\author{
Ali Parviz\textsuperscript{1,2} \thanks{ This work was done when the author was visiting the National Institute of Informatics (NII).} \quad \textbf{Yuichi  Yoshida}\textsuperscript{3} \thanks{Supported by JSPS KAKENHI Grant Number 20H05965} \\
\textsuperscript{1}Mila - Quebec AI Institute,
\textsuperscript{2}New Jersey Institute of Technology,
\textsuperscript{3}National Institute of Informatics
\\
}

%% file: chapters/00_abstract.tex
\begin{abstract}
While signed-directed graphs have been studied using linear Laplacians in the design of graph neural networks, relatively little research has focused on developing non-linear Laplacian operators for such networks. We introduce a non-linear Laplacian operator specific to signed and directed networks (NLSD). This non-linear operator extends the concepts of the signed Laplacian for signed graphs and the Laplacian for directed graphs. The NLSD calculates node-specific potentials based on features More precisely, if the potential discrepancy is not aligned with the edge direction, we ignore it (and vice versa) leveraging message-passing techniques only across edges where potential discrepancies align with the edge's direction. Utilizing this novel operator, we propose an efficient spectral GNN framework (NLSD-GNN). We conducted comprehensive evaluations focusing on node classification and link prediction, examining scenarios involving signed, directional, or both types of information. Our findings reveal that this spectral GNN framework not only integrates signed and directional data effectively but also achieves superior performance across diverse datasets.

\end{abstract}

%% file: chapters/01_Introduction.tex
\section{Introduction}

Multiple real-life systems entail interactions that may be positive or negative and not always symmetrical~\cite{he2024pytorch}. Such systems are best represented as undirected or directed graphs with {\em signed} edges, i.e.~edges with an additional (+/-) sign attribute. In social networks, for example, users may share or oppose political views, trust or distrust recommendations from others, or harbor feelings of liking or disliking toward each other~\cite{kumar2016edge,huang2021sdgnn}. These diverse relationships can be effectively represented by signed graphs, where edges are assigned either positive or negative values to reflect affinity or conflict, and directional information could model the influence of one person on another~\cite{ju2020new}.
Signed-directed networks are also prevalent in contexts such as analyzing time-series data~\cite{bennett2022detection}, identifying influential groups within social networks, and generating rankings through pairwise comparisons~\cite{he2022gnnrank}. Moreover, these networks provide an intuitive framework for analyzing group conflicts and enhancing positive influence during opinion formation~\cite{zheng2015signed}.

Not surprisingly, directed and signed graphs have attracted significant attention in graph representation learning. Graph Neural Networks (GNNs) in this context mimic standard GNNs and employ message passing, or equivalently, local aggregation of node features. The operators used by standard GNNs have well-understood spectral properties, which when seen globally relate to the combinatorial structure of the graph. However, the generalization of these operators to the signed-directed setting is far from obvious; indeed much of the existing work revolves around definitions of various {\em linear operators} for implementing message-passing. 
Table~\ref{table:lap_tbl} summarizes the important aspects of these different Laplacians. Our work relies on a fragment of spectral theory that defines {\em nonlinear} Laplacian operators to encode directed graphs, and proves their expressivity in capturing -via Cheeger inequalities- combinatorial properties of the underlying directed graph~\cite{Yoshida2016NonlinearLF}. More specifically, these Laplacians can be viewed as functions of a feature vector $x$, that combine two steps: (i) Edge dropouts that enforce message-passing only along directed edges $(u,v)$ where $x_u\geq x_v$, (ii) The action of a symmetrized Laplacian on the subgraph induced by the edge dropouts. We begin by extending the definition of~\cite{Yoshida2016NonlinearLF} and propose a nonlinear Laplacian that takes into account both edge direction and sign. While the extension is natural, it has not been discussed before.


The theoretical appeal and applied value of these nonlinear Laplacians motivate two questions:  
\textbf{Q1:} Can these nonlinear Laplacians be extended to directed graphs with additional edge signs?  
\textbf{Q2:} Can they be leveraged in the context of graph learning?  
We provide positive answers to both of these questions.
We begin by extending the definition of~\cite{Yoshida2016NonlinearLF} and propose a non-linear Laplacian that takes into account both edge direction and sign. While the extension is natural, it has not been discussed before. Using these nonlinear Laplacians in the context of GNNs, while simultaneously aiming for computational efficiency, is a trickier problem. In our method NLSD-GNN, we introduce a module that projects the node feature vectors $X \in {\mathbb R}^{n\times d}$ to scalar values. The module can be a fixed random projection (for efficiency), or a parametric trainable function, and it can be viewed as a function that assigns {\em potentials} to the nodes of the graph. Then, these potentials values are used to select a subgraph of the input graph, induced by the signed-directed edges. The subgraph is symmetrized into the signed-undirected counterpart and self-loops are added adequately. Then, the standard {\em linear} operator of the signed-undirected graph is used for the message propagation. 
Experimental validation shows that our methods improve upon the best known results on several benchmarks for transductive node classification and link prediction.

%% file: chapters/03_Spectheory.tex
\section{Preliminary}



Let $G=(V,E,w)$ be a (weighted) directed graph, where \( V \) represents the set of vertices, \( E \) is the set of directed edges, and \( w: E \rightarrow \mathbb{R}_{\geq 0} \) is the weight function.
Although self-loops are allowed, we assume there are no parallel edges.
For a vertex $v \in V$, the (weighted) \emph{indegree} and \emph{outdegree} of $v$ is defined to be $d^-(v) := \sum_{e=(u,v)\in E}w(e)$ and $d^+(v)=\sum_{e=(v,u)\in E}w(e)$, respectively.
The \emph{total degree} of $v$ is $d(v) := d^-(v) + d^+(v)$, which is the total weight of edges incident to $v$.

A signed graph $G=(V,E^+, E^-,w)$ consists of a vertex set $V$, two sets $E^+,E^-$ of undirected edges, and a weight function $w:E^+ \cup E^- \to \mathbb{R}_{\geq 0}$.
Similarly, a \emph{signed-directed graph} $G=(V,E^+, E^-,w)$ consists of a vertex set $V$, two sets $E^+,E^-$ of directed edges, and a weight function $w:E^+ \cup E^- \to \mathbb{R}_{\geq 0}$.
We call edges in $E^+$ and $E^-$ \emph{positive edges} and \emph{negative edges}, respectively.
The \emph{total degree} of $v$ is the total weight of edges incident to $v$, including both positive and negative ones. We discussed the related in the Appendix~\ref{related}

\section{Non Linear Signed-Directed
Laplacian} 
\label{sec:specdir}

We review the existing Laplacians for undirected, directed, signed and propose a novel non-linear Laplacian for signed-directed graphs.
\paragraph{Laplacian for undirected graphs.}
Let $G=(V,E,w)$ be a weighted undirected graph.
The \emph{degree matrix} of $G$ is the diagonal matrix $D_G \in \mathbb{R}^{V \times V}$, where $(D_G)_{vv}$ is the (weighted) degree of $v\in V$, i.e., $\sum_{e\in E: v \in e}w(e)$.
The \emph{adjacency matrix} of $G$ is a matrix $A_G \in \mathbb{R}^{V \times V}$, where $(A_G)_{uv}=(A_G)_{vu} = w(\{u,v\})$ if $\{u,v\} \in E$ and $(A_G)_{uv}=(A_G)_{vu}=0$ otherwise.
The \emph{Laplacian} $L_G \in \mathbb{R}^{V\times V}$ of $G$ is defined to be $D_G - A_G$.
It is well known that its quadratic form satisfies $Q_G(x) := x^\top L_G x = \sum_{e = \{u,v\} \in E}w(e) (x_u-x_v)^2$, and hence for a set $S\subseteq V$, $Q(\bm{1}_S)$ gives the cut weight of the vertex set $S$, where $\bm{1}_S \in \mathbb{R}^V$ is the characteristic vector of $S$.
The normalized Laplacian $\mathcal{L}_G$ is defined as $\mathcal{L}_G := D_G^{-1/2}L_G D_G^{-1/2}$.

\paragraph{Laplacian for directed graphs.}
Here, we define nonlinear Laplacian for directed graphs~\cite{Yoshida2016NonlinearLF}:
\begin{definition}\label{def:directed-laplacian}
Consider a directed graph $G = (V, E, w)$. 
The \emph{Laplacian} $L_G: \mathbb{R}^V \rightarrow \mathbb{R}^V$ is defined to output a vector $L_G(x)$ for a vector $x \in \mathbb{R}^V$ via the following procedure:
\begin{enumerate}
    \itemsep=0pt
    \item Construct an undirected graph $G_x$ on the vertex set $V$ as follows: For each directed edge $e = (u, v) \in E$, include an undirected edge $\{u, v\}$ of weight $w(e)$ in $G_x$ if $x_u \geq x_v$, and introduce self-loops $\{u, u\}$ and $\{v, v\}$ of weight $w(e)/2$ otherwise.
    \item Let $L_{G_x} \in \mathbb{R}^{V\times V}$ be the Laplacian matrix for the undirected graph $G_x$. 
    \item The vector $L_G(x)$ is then derived as $L_{G_x} x$.
\end{enumerate}
The resulting undirected graph $G_x$ and the corresponding $L_G(x)$
are shown in Figure~\ref{fig:directed}. We note that the degree of any vertex $v \in V$ in $G_x$ is exactly equal to the total degree of $v$ in $G$.
\end{definition}


An important feature of the Laplacian for directed graphs is that its quadratic form satisfies $Q_G(x) := x^\top L_G x = \sum_{e = \{u,v\} \in E}w(e) \max\{x_u-x_v,0\}^2$~\cite{Yoshida2016NonlinearLF}.
Hence for a set $S\subseteq V$, $Q(\bm{1}_S)$ gives the cut weight of the vertex set $S$.

\paragraph{Laplacian for signed graphs.}

Let $G=(V,E^+,E^-,w)$ be a signed graph, and let $G^+ = (V,E^+,w)$ and $G^-=(V,E^-,w)$ be the undirected graphs consisting of positive and negative, respectively, edges.
Then, let $L_G^+ \in \mathbb{R}^{V \times V}$ be the Laplacian of $G^+$, and $L_G^- \in \mathbb{R}^{V \times V}$ be the matrix defined as $L_G^- := D_{G^-}+A_{G^-} = 2D_{G^-} - L_{G^-}$.
Then, the Laplacian $L_G \in \mathbb{R}^{V \times V}$ of $G$ is $L_G = L_G^+ + L_G^-$~\cite{hou2003laplacian}.
We can observe that its quadratic form $Q_G(x) = x^\top L_G x$ can be written as 
\begin{equation*}
\resizebox{\columnwidth}{!}{
    $Q_G(x) = \sum_{e=\{u,v\} \in E^+} w(e)(x_u - x_v)^2 + \sum_{e=\{u,v\} \in E^-} w(e)(x_u + x_v)^2$
}
\end{equation*}

Note that the first term is equal to the quadratic form of the Laplacian of $G^+$.
The second term becomes small when the values for the endpoints of an edge have  opposite signs.
For $S\subseteq V$, $Q_G(\bm{1}_{S} - \bm{1}_{V\setminus S})$ is four times the total weight of positive edges cut by $S$ plus the total weight of negative edges uncut by $S$.

\subsection{Laplacian for Signed-Directed Graphs}\label{subsec:signed-directed-laplacian}

\begin{figure}[htbp!]
\tiny
    \centering
    \begin{subfigure}[b]{0.40\textwidth}
        \centering
        \includegraphics[width=\textwidth]{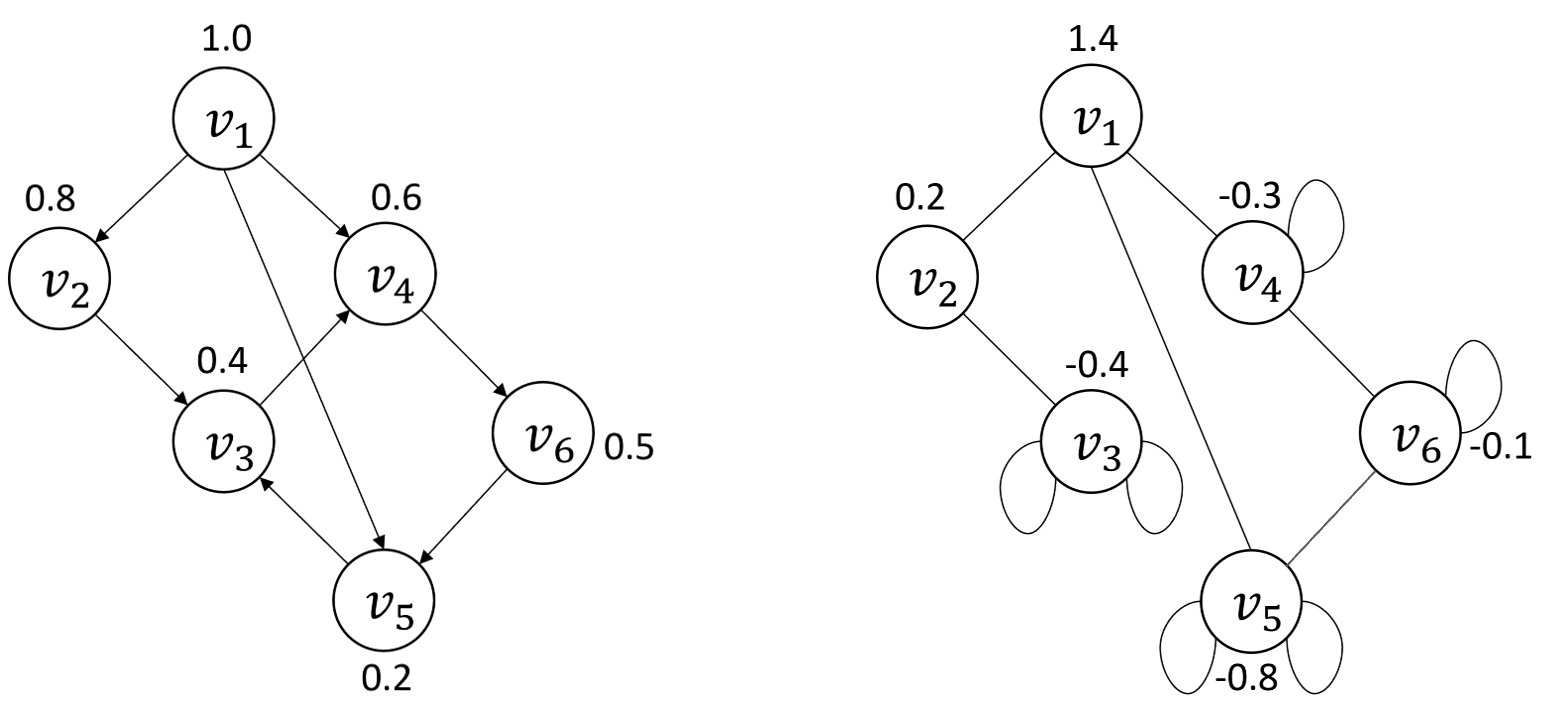}
        \caption{Directed graph}\label{fig:directed}
    \end{subfigure}
    \vspace{1mm}
    \begin{subfigure}[b]{0.34\textwidth}
        \centering
        \includegraphics[width=\textwidth]{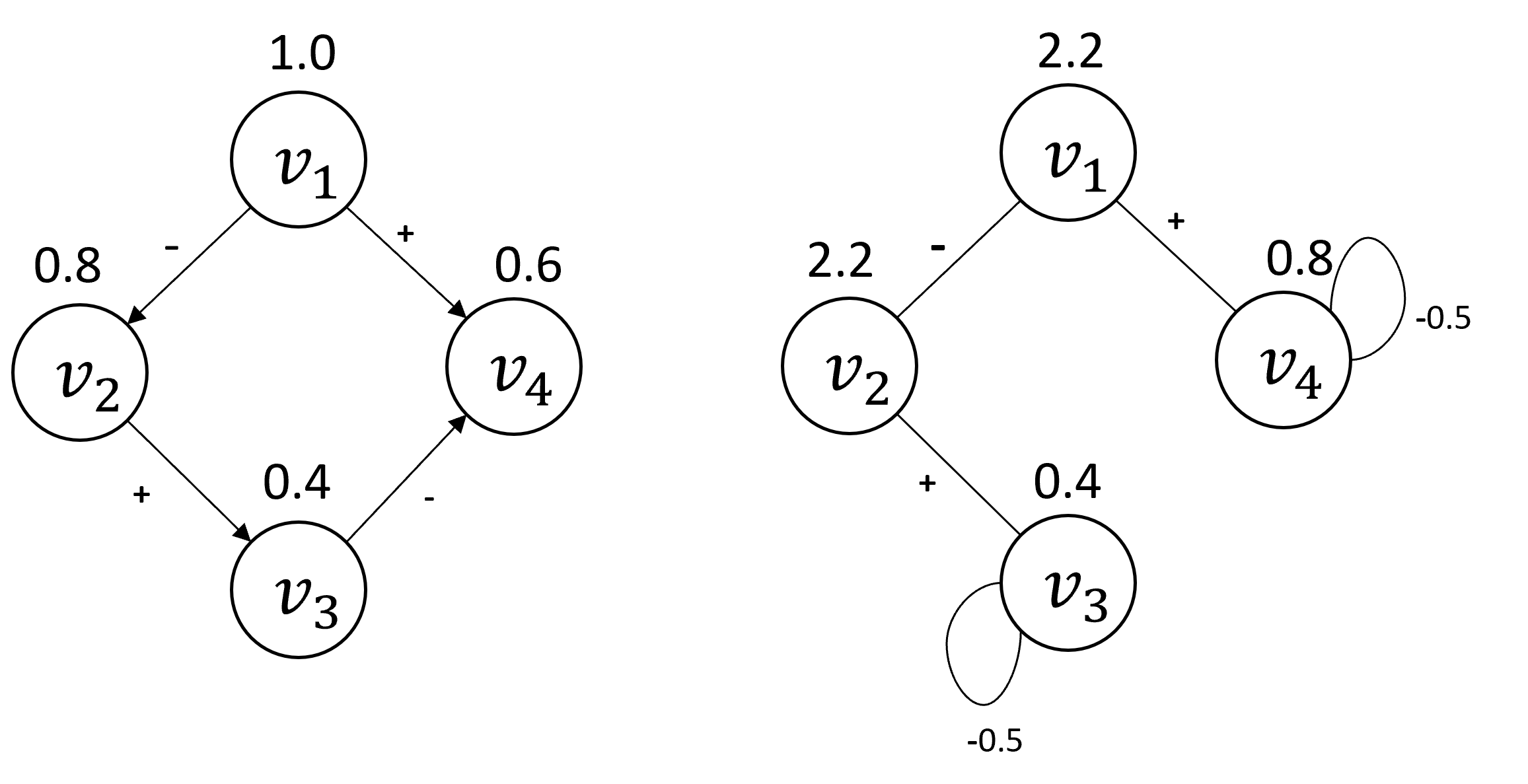}
        \caption{Signed-directed graph }\label{fig:signed-directed}
    \end{subfigure}
    \caption{The left figures show unsigned (\ref{fig:directed}) and signed (\ref{fig:signed-directed}) directed graphs $G$ and vectors $x$ while the right figures show undirected graphs $G_x$ and vectors $L_G(x)$.
    }
    \label{fig:graph}
\end{figure}

We define nonlinear Laplacian for signed-directed graphs by naturally unifying Laplacian for directed graphs and that for signed graphs:
\begin{definition}\label{def:signed-directed-laplacian}
Consider a signed-directed graph $G = (V,E^+\cup E^-,w)$. 
The \emph{Laplacian} $L_G: \mathbb{R}^V \rightarrow \mathbb{R}^V$ is defined to output a vector $L_G(x)$ for a vector $x \in \mathbb{R}^V$ via the subsequent procedure:
\begin{enumerate}
    \itemsep=0pt
    \item Construct a signed-undirected graph $G_x$ on the vertex set $V$ as follows: 
    For each edge $(u, v) \in E^+ \cup E^-$, include an undirected edge of the same sign $e = \{u, v\}$ in $G_x$ of weight $w(e)$ if $x_u \geq x_v$, and introduce self-loops of the same sign $\{u, u\}$ and $\{v, v\}$ of weight $w(e)/2$ otherwise. 
    \item Let $L_{G_x} \in \mathbb{R}^{V \times V}$ be the Laplacian matrix for the signed-undirected graph $G_x$. 
    \item The vector $L_G(x)$ is then derived as $L_{G_x} x$.
\end{enumerate}
\end{definition}
Figure~\ref{fig:signed-directed} shows an example of a signed-directed graph, a vector $x \in \mathbb{R}^V$, and the computation of the signed-undirected graph $G_x$ and the corresponding $L_G(x)$.

The \emph{normalized Laplacian} $\mathcal{L}_G: \mathbb{R}^V \to \mathbb{R}^V$ of $G$ is defined as $\mathcal{L}_G: x \mapsto D_G^{-1/2}L_G (D_G^{-1/2} x)$, where $D_G \in \mathbb{R}^{V \times V}$ is the diagonal matrix with total degree on the diagonal entries. 
For a signed-directed graph $G=(V,E^+,E^-,w)$, let $Q_G(x) = x^\top L_G(x)$ be its quadratic form.
Then, we have 
\begin{multline*}
    Q_G(x) = x^\top L_{G_x}^+ x + x^\top L_{G_x}^- x \\
    = x^\top L_{G^+_x} x + x^\top (2D_{G^-_x} - L_{G^-_x}) x \\
    = x^\top L_{G^+_x} x + x^\top (2D_{G^-} - L_{G^-_x}) x \\
    = \sum_{e = (u,v) \in E^+}w(e) \max\{x_u - x_v,0\}^2 \\
    + \sum_{e = (u,v) \in E^-}w(e) (2x_u^2 + 2x_v^2 -  \max\{x_u - x_v,0\}^2).
\end{multline*}
The second term can be written as
\[
    \sum_{e=(u,v)\in E^-}w(e) \cdot 
    \begin{cases}
        (x_u+x_v)^2 & \text{if }x_u \geq x_v\\
        2x_u^2 + 2x_v^2 & \text{otherwise}
    \end{cases}
\]
For $S\subseteq V$, $Q_G(\bm{1}_S - \bm{1}_{V\setminus S})$ is four times the total weight of positive edges cut by $S$ plus the total weight of negative edges $(u,v) \in E^-$ such that $x_u \leq x_v$.
Hence to minimize $Q_G(x)$, we want to embed vertices onto a one-dimensional line so that the tail vertices are to the left of the head vertices for positive edges and the tail vertices are to the right of the head vertices for negative edges.

When $\lambda \in \mathbb{R}$ and $v \in \mathbb{R}^V$ with $v\neq \bm{0}$ satisfy $L_G(v) = \lambda v$, we say that $\lambda$ is an \emph{eigenvalue} of $L_G$ and that $v$ is a corresponding eigenvector. The following theorems show that $L_G$ and $\mathcal{L}_G$ satisfy properties analogous to the
traditional graph Laplacians. The proofs are in Appendix B. 
\begin{theorem}\label{thm:psd-and-bounded-eigenvalues}
    For any signed directed graph G, The operators $L_G$ and $\mathcal{L}_G$ are positive semidefinite, i.e., all the eigenvalues are nonnegative.
\end{theorem}
\begin{theorem}\label{thm:psd-and-bounded-eigenvalues}
     For any signed directed graph G, all the eigenvalues of $\mathcal{L}_G$ are in the range $[0,2]$.
\end{theorem}

%% file: chapters/04_orgGNN.tex
\section{NLSD-GNN architecture}\label{sec:NLSD-GNN}

\begin{figure*}[t]
  \centering
  \includegraphics[width=0.85\textwidth, height=0.30\textwidth]{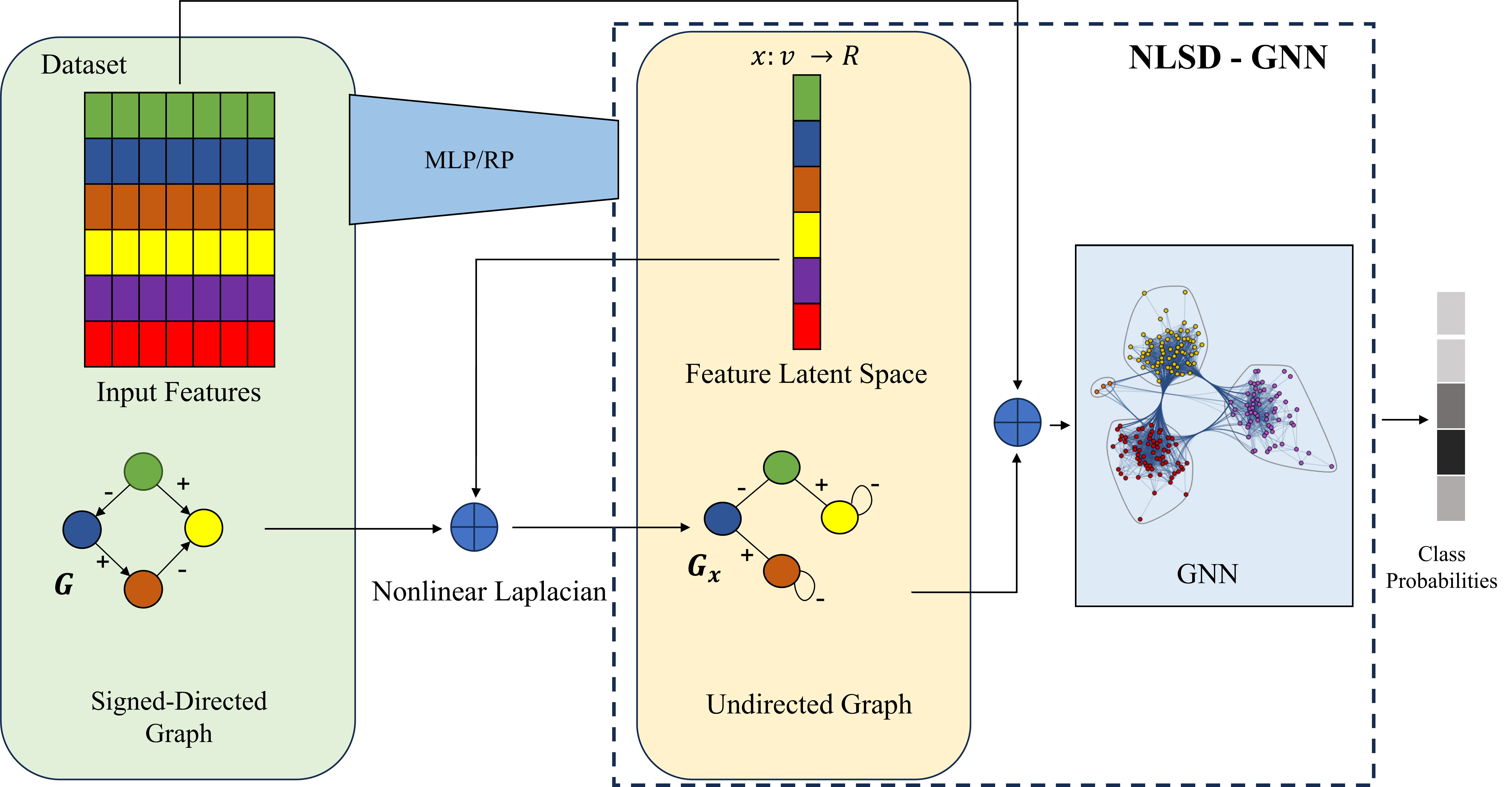}
  \caption{
  Overview of our workflow, NLSD-GNN, with a toy example. In the first block, we use an MLP or random projection to generate the feature latent space. In the next step, by using the nonlinear Laplacian transformation we calculate the counterpart signed-undirected graph by using the feature representation. The last block of the pipeline consists of passing the input features and the counterpart graph to spectral GNN layers to get the final prediction for the downstream task.  }

  \label{fig:pipleline}
\end{figure*}

In this section, we show how to use a normalized or unnormalized nonlinear Laplacian introduced in \Cref{def:signed-directed-laplacian}, to define convolution on a
signed-directed. 

\begin{algorithm}
\caption{Algorithm for NLSD-GNN-Fast}
\label{alg:NLSD-GNN-Fast}
\small
\textbf{Input:} Signed-directed graph $G = (V, E^+ \cup E^-, w)$, features $H \in \mathbb{R}^{N \times d}$, labeled vertices $V_L$ \\
\textbf{Output:} Labels for all nodes in $V - V_L$

\begin{algorithmic}[1] 
    \STATE Apply random projection $x = H \cdot r$ where $r \in \mathbb{R}^{d \times 1}$
    
    \STATE \textbf{Graph Transformation:}
    \STATE Construct signed-undirected graph $G_x$ on vertex set $V$
    \FOR{each edge $(u, v)$ in $E^+ \cup E^-$}
        \IF{$x_u \geq x_v$}
            \STATE Include edge $e = \{u, v\}$ in $G_x$ with weight $w(e)$
        \ELSE
            \STATE Introduce self-loops of the same sign $\{u, u\}$ and $\{v, v\}$ of weight $w(e)/2$
        \ENDIF
    \ENDFOR
    \STATE Set $\tilde{A}_{G_x}$ as the adjacency matrix of $G_x$
    
    \FOR{each epoch $\tau$}
        \FOR{layer $\ell = 1$ to $2$}
            \STATE Compute $Z = \text{softmax}(\tilde{A}_{G_x} \text{ReLU}(\tilde{A}_{G_x} H \Theta^{(1)}) \Theta^{(2)})$
            \STATE Update parameters $\Theta_\tau$ to minimize cross-entropy loss
        \ENDFOR
        \STATE Label nodes in $V - V_L$ using $Z$
    \ENDFOR
\end{algorithmic}
\end{algorithm}

\subsection{Spectral Convolution on Graphs.}
Signal on graphs can be filtered using the eigendecomposition of graph Laplacian: $\mathcal{L} = U\Lambda U^T,$ where  $U$  contains all of the eigenvectors of  $\mathcal{L}$ and $ \Lambda = \text{diag}(\lambda_1, \ldots, \lambda_n)$  is a diagonal matrix consisting of the corresponding eigenvalues as its diagonal. 
Given a signal $h$ , the graph Fourier transform of $h$  can be defined as 
$\hat{h} = U^T h,$ and its inverse operation as $h = U\hat{h}.$
The spectral convolution filter $\mathcal{F}$ can be formulated as: $\mathcal{F} \ast h = U f(\Lambda) U^T h$,
where $f(\Lambda)$  is the graph convolution filter function corresponding to the signal frequency  $\Lambda$ . By defining the graph convolution kernel as:
\begin{align}
    \mathcal{F} = Uf(\Lambda)U^T,
\end{align}
The following formulation connects the spectral-based and spatial-based GNNs in a unified form~\cite{Balcilar2020BridgingTG}:
\begin{align}
\label{eq:convequ1}
  H^{(l+1)} = \sigma \left( \sum_{k=1}^K \mathcal{F}_k H^{(l)} W^{(l,k)} \right),  
\end{align}

where $\mathcal{F}_k$  denotes the k-th graph convolution kernel, $\sigma $ denotes the nonlinear activation function such as ReLU,  $H^{(l)}$  is the node representations at  $l$-th layer, and  $W^{(l,k)}$  is the learnable parameter matrix.
\subsection{The NLSD-GNN architecture}\label{subsec:NLSD-GNN}

We now introduce our architecture, NLSD-GNN.
At the $\ell$-th layer, each node is associated with a $F_\ell$-dimensional feature vector in a graph, and the feature \emph{matrix} for all nodes is represented as $H^{(\ell)} \in \mathbb{R}^{N \times F_\ell}$. 
Following the Laplacian construction steps outlined in Definition 2, we require a \emph{vector} $x \in \mathbb{R}^N$ to constructing an undirected counterpart of the graph.
To effectively reduce the dimensionality of our node feature matrix $H^{(\ell)}$, we consider the following two schemes:
\paragraph{NLSD-GNN-MLP.} We employ a simple MLP architecture with a tanh as an activation function between the layers except for the last layer. At each layer, we progressively halve the dimensionality until reaching dimensionality of one. Now, we can map the original directed graph to its undirected counterpart by leveraging this one-dimensional feature vector. \Cref{fig:pipleline} illustrates our architecture.
\paragraph{NLSD-GNN-Fast.} As a faster alternative, we also propose an architecture inspired by~\cite{Louis2014HypergraphMO, Yadati2018HyperGCNAN}, which creates the feature vector by leveraging a simple random projection.

Let $G=(V,E^+,E^-,w)$ be a signed-directed graph.
Then, we train our GNN using the original feature matrix and the nonlinear Laplacian graph generated in the previous step as inputs.
Following \Cref{eq:convequ1}, to generate the $\ell$-th layer from the $(\ell - 1)$-th layer, we compute the matrix $H^{(\ell)}$ as follows:
\begin{align}
\label{eq:convequ2}
H^{(\ell)} = \sigma \left( \tilde{D}_G^{-\frac{1}{2}} \tilde{A}_{G_x} \tilde{D}_G^{-\frac{1}{2}}  H^{(\ell-1)} \Theta^{(\ell)} \right),
\end{align}
where $x \in \mathbb{R}^N$ is obtained via an MLP for NLSD-GNN-MLP and a random projection for NLSD-GNN-Fast. Following the convolutional layers, we transform the matrix $H^{(L)} \in \mathbb{R}^{N \times F_L}$ into $N \times n_c$ matrix by applying a linear layer and then apply a weight matrix $W^{(L+1)} \in \mathbb{R}^{F_L \times n_c}$ from the right, where $n_c$ is the number of classes.
Finally, we apply softmax on the output of the final layer. In our experiments, we set $L$ to 2 or 3.
For link prediction, we apply the same method up to the last layer, then concatenate the rows associated with pairs of nodes to generate the edge features. we discussed the complexity of our method in Appendix.

%% file: chapters/05_Experiments.tex
\section{Experiments}\label{sec:experiments}




\begin{table*}[t!]
\small
 \centering
 \caption{Node classification accuracy ($\%$) and its standard deviation for directed graphs. Top three models are color-coded  as \textcolor{OliveGreen}{First}, \textcolor{BlueViolet}{Second}, \textcolor{Maroon}{Third}.}  \label{tab:result1}
\begin{tabular}{lllllll}
    \toprule
     Type &Method &Cornell &Texas& Wisconsin &Cora-ML& CiteSeer  \\
     \midrule
    \multirow{2}{*}
    {Spectral}& GCN &59.0 \tiny{$\pm$ }6.4 &58.7 \tiny{$\pm$ }3.8 & 55.9 \tiny{$\pm$ }5.4 &82.0 \tiny{$\pm$ }1.1 & 66.0 \tiny{$\pm$ }1.5 \\
    &ChebNet& \textcolor{Maroon}{76.3 \tiny{$\pm$ }3.2}& 79.2 \tiny{$\pm$ }7.5 &81.6 \tiny{$\pm$ }6.3 &80.0 \tiny{$\pm$ }1.8& 66.7 \tiny{$\pm$ }1.6 \\
    
\midrule
\multirow{4}{*}{Spatial}& APPNP &58.7 \tiny{$\pm$ }4.0 &57.0 \tiny{$\pm$ }4.8& 51.8 \tiny{$\pm$ }7.4& \textcolor{OliveGreen}{82.6 \tiny{$\pm$ }1.4}& 66.9 \tiny{$\pm$ }1.8 \\
&SAGE &\textcolor{OliveGreen}{80.0 \tiny{$\pm$ }6.1}& \textcolor{Maroon}{84.3 \tiny{$\pm$ }5.5}& 83.1 \tiny{$\pm$ }4.8& \textcolor{BlueViolet}{82.3 
 \tiny{$\pm$ }1.2}& 66.0 \tiny{$\pm$ }1.5\\
&GIN& 57.9 \tiny{$\pm$ }5.7 &65.2 \tiny{$\pm$ }6.5& 58.2 \tiny{$\pm$ }5.1 &78.1 \tiny{$\pm$ }2.0& 63.3 \tiny{$\pm$ }2.5\\
&GAT& 57.6 \tiny{$\pm$ }4.9& 61.1 \tiny{$\pm$ }5.0 &54.1 \tiny{$\pm$ }4.2& 81.9 \tiny{$\pm$ }1.0& 67.3 \tiny{$\pm$ }1.3\\
 \midrule
 \multirow{4}{*}{Directed}&DGCN& 67.3 \tiny{$\pm$ }4.3& 71.7 \tiny{$\pm$ }7.4& 65.5 \tiny{$\pm$ }4.7 &81.3 \tiny{$\pm$ }1.4 &66.3 \tiny{$\pm$ }2.0 \\
 &Digraph &66.8 \tiny{$\pm$ }6.2& 64.9 \tiny{$\pm$ }8.1& 59.6 \tiny{$\pm$ }3.8& 79.4 \tiny{$\pm$ }1.8& 62.6 \tiny{$\pm$ }2.2\\
 &DiGraphIB &64.4 \tiny{$\pm$ }9.0& 64.9 \tiny{$\pm$ }13.7& 64.1 \tiny{$\pm$ }7.0& 79.3 \tiny{$\pm$ }1.2 &61.1 \tiny{$\pm$ }1.7\\
 &MagNet & 74.0 \tiny{$\pm$ }4.5& 83.3 \tiny{$\pm$ }6.1& \textcolor{Maroon}{85.7 \tiny{$\pm$ }3.2}& 79.8 \tiny{$\pm$ }2.5& \textcolor{Maroon}{67.5 \tiny{$\pm$ }1.8} \\
  \midrule
  \multicolumn{2}{l}{\textbf{NLSD-GNN-MLP}} & \textcolor{BlueViolet}{78.1 \tiny{$\pm$ }4.5}& \textcolor{OliveGreen}{85.5 \tiny{$\pm$ }5.2}& \textcolor{BlueViolet}{86.8 \tiny{$\pm$ }4.3}& 79.3 \tiny{$\pm$ }1.2& \textcolor{BlueViolet}{67.6 \tiny{$\pm$ }2.6} \\ \cline{3-7}
  \multicolumn{2}{l}{\textbf{NLSD-GNN-Fast}} & 77.8 \tiny{$\pm$ }3.9& \textcolor{BlueViolet}{85.4 \tiny{$\pm$ }4.2}& \textcolor{OliveGreen}{87.2 \tiny{$\pm$ }3.5}  & \textcolor{Maroon}{82.2 \tiny{$\pm$ }2.5}& \textcolor{OliveGreen}{68.3 \tiny{$\pm$ }1.2}  \\
    \bottomrule
\end{tabular}
\end{table*}

\begin{table*}[t!]
\scriptsize
\centering
\caption{Test accuracy (\%) comparison the signed-directed link prediction. Top three models are color-coded as \textcolor{OliveGreen}{First}, \textcolor{BlueViolet}{Second}. } \label{tab:result3}
\label{tab:datasets_performance}
\begin{tabular}{@{}llcccccccc@{}}
\toprule
{Dataset} & {Task} & {SGCN} & {SDGNN} & {SiGAT} & {SNEA} & {SSSNET} & {SigMaNet} & {MSGNN} & {NLSD-GNN} \\ \midrule
\multicolumn{1}{c}{} & SP & 64.7$\pm$0.9 & 64.5$\pm$1.1 & 62.9$\pm$0.9 & 64.1$\pm$1.3 & 67.4$\pm$1.1 & 47.8$\pm$3.9 & \textcolor{BlueViolet}{71.3$\pm$1.2} & \textcolor{OliveGreen}{72.7$\pm$0.8} \\
\multicolumn{1}{c}{} & DP & 60.4$\pm$1.7 & 61.5$\pm$1.0 & 61.9$\pm$1.9 & 60.9$\pm$1.7 & 68.1$\pm$2.3 & 49.4$\pm$3.1 & \textcolor{OliveGreen}{72.5$\pm$1.5} & \textcolor{BlueViolet}{70.5$\pm$0.5} \\
\multicolumn{1}{c}{BitCoin-Alpha} & 3C & 81.4$\pm$0.5 & 79.2$\pm$0.9 & 77.1$\pm$0.7 & 83.2$\pm$0.5 & 78.3$\pm$4.7 & 37.4$\pm$16.7 & \textcolor{BlueViolet}{84.4$\pm$0.6} & \textcolor{OliveGreen}{85.6$\pm$1.6} \\
\multicolumn{1}{c}{} & 4C & 51.1$\pm$0.8 & 52.5$\pm$1.1 & 49.3$\pm$0.7 & 52.4$\pm$1.8 & 54.3$\pm$2.9 & 20.6$\pm$6.3 & \textcolor{BlueViolet}{58.5$\pm$0.7} & \textcolor{OliveGreen}{58.7$\pm$0.3} \\
\multicolumn{1}{c}{} & 5C & 79.5$\pm$0.3 & 78.2$\pm$0.5 & 76.5$\pm$0.3 & 81.1$\pm$0.3 & 77.9$\pm$0.3 & 34.2$\pm$6.5 & \textcolor{BlueViolet}{81.9$\pm$0.9} & \textcolor{OliveGreen}{82.6$\pm$0.3} \\
\hline
\addlinespace
\multicolumn{1}{c}{} & SP & 65.6$\pm$0.9 & 65.3$\pm$1.2 & 62.8$\pm$1.3 & 67.7$\pm$0.5 & 70.1$\pm$1.2 & 50.0$\pm$2.3 & \textcolor{BlueViolet}{73.0$\pm$1.4} & \textcolor{OliveGreen}{73.8$\pm$0.4} \\
\multicolumn{1}{c}{} & DP & 63.8$\pm$1.2 & 63.2$\pm$1.5 & 64.0$\pm$2.0 & 65.3$\pm$1.2 & 69.6$\pm$1.0 & 48.4$\pm$4.9 & \textcolor{OliveGreen}{71.8$\pm$1.1} & \textcolor{OliveGreen}{71.8$\pm$0.7} \\
\multicolumn{1}{c}{BitCoin-OTC} & 3C & 79.0$\pm$0.7 & 77.3$\pm$0.7 & 73.6$\pm$0.7 & 82.2$\pm$0.4 & 76.9$\pm$1.1 & 26.8$\pm$10.9 & \textcolor{BlueViolet}{83.3$\pm$0.7} & \textcolor{OliveGreen}{84.2$\pm$1.7} \\
\multicolumn{1}{c}{} & 4C & 51.5$\pm$0.4 & 55.3$\pm$0.8 & 51.2$\pm$1.8 & 56.9$\pm$0.7 & 57.0$\pm$2.0 & 23.3$\pm$7.4 & \textcolor{BlueViolet}{59.8$\pm$0.7} & \textcolor{OliveGreen}{60.5$\pm$0.8} \\
\multicolumn{1}{c}{} & 5C & 77.4$\pm$0.7 & 77.3$\pm$0.8 & 74.1$\pm$0.5 & 80.5$\pm$0.5 & 74.0$\pm$1.6 & 25.9$\pm$6.2 & \textcolor{BlueViolet}{80.9$\pm$0.9} & \textcolor{OliveGreen}{82.5$\pm$0.3} \\
\hline
\addlinespace
\multicolumn{1}{c}{} & SP & 74.7$\pm$0.5 & 74.1$\pm$0.7 & 64.0$\pm$1.3 & 70.6$\pm$1.0 & 86.6$\pm$2.2 & 57.9$\pm$5.3 & \textcolor{BlueViolet}{92.4$\pm$0.2} & \textcolor{OliveGreen}{93.2$\pm$0.6} \\
\multicolumn{1}{c}{} & DP & 74.8$\pm$0.9 & 74.2$\pm$1.4 & 62.8$\pm$0.9 & 71.1$\pm$1.1 & 87.8$\pm$1.0 & 53.0$\pm$4.0 & \textcolor{BlueViolet}{93.1$\pm$0.1} & \textcolor{OliveGreen}{93.3$\pm$0.1} \\
\multicolumn{1}{c}{Slashdot} & 3C & 69.7$\pm$0.3 & 66.3$\pm$1.8 & 49.1$\pm$1.2 & 72.5$\pm$0.7 & 79.3$\pm$1.2 & 42.0$\pm$7.9 & \textcolor{BlueViolet}{86.1$\pm$0.3} & \textcolor{OliveGreen}{86.8$\pm$0.5} \\
\multicolumn{1}{c}{} & 4C & 63.2$\pm$0.3 & 64.0$\pm$0.7 & 53.4$\pm$0.2 & 60.5$\pm$0.6 & 72.7$\pm$0.6 & 25.7$\pm$8.9 & \textcolor{BlueViolet}{78.2$\pm$0.3} & \textcolor{OliveGreen}{79.1$\pm$1.3} \\
\multicolumn{1}{c}{} & 5C & 64.4$\pm$0.3 & 62.6$\pm$2.0 & 44.4$\pm$1.4 & 66.4$\pm$0.5 & 70.4$\pm$0.7 & 19.3$\pm$8.6 & \textcolor{BlueViolet}{76.8$\pm$0.6} & \textcolor{OliveGreen}{78.4$\pm$0.3} \\
\hline
\addlinespace
\multicolumn{1}{c}{} & SP & 62.9$\pm$0.5 & 67.7$\pm$0.8 & 63.6$\pm$0.5 & 66.5$\pm$1.0 & 78.5$\pm$2.1 & 53.3$\pm$10.6 & \textcolor{BlueViolet}{85.4$\pm$0.5} & \textcolor{OliveGreen}{86.1$\pm$0.4} \\
\multicolumn{1}{c}{} & DP & 61.7$\pm$0.5 & 67.9$\pm$0.6 & 63.6$\pm$0.8 & 66.4$\pm$1.2 & 73.9$\pm$6.2 & 49.0$\pm$3.2 & \textcolor{BlueViolet}{86.3$\pm$0.3} & \textcolor{OliveGreen}{86.7$\pm$0.6} \\
\multicolumn{1}{c}{Epinions} & 3C & 70.3$\pm$0.8 & 73.2$\pm$0.8 & 52.3$\pm$1.3 & 72.8$\pm$0.2 & 72.7$\pm$2.0 & 30.5$\pm$8.3 & \textcolor{BlueViolet}{83.1$\pm$0.5} & \textcolor{OliveGreen}{85.2$\pm$0.7} \\
\multicolumn{1}{c}{} & 4C & 66.7$\pm$1.2 & 71.0$\pm$0.6 & 62.3$\pm$0.5 & 69.5$\pm$0.7 & 70.2$\pm$5.2 & 29.9$\pm$6.4 & \textcolor{BlueViolet}{78.7$\pm$0.9} & \textcolor{OliveGreen}{79.7$\pm$0.1} \\
\multicolumn{1}{c}{} & 5C & 73.5$\pm$0.8 & 76.6$\pm$0.7 & 52.9$\pm$0.7 & 74.2$\pm$0.1 & 70.3$\pm$4.6 & 22.1$\pm$6.1 & \textcolor{BlueViolet}{80.5$\pm$0.5} & \textcolor{OliveGreen}{81.3$\pm$0.1} \\
\bottomrule
\end{tabular}
\end{table*}

The node classification task consists in predicting the class label to which each node belongs. For this task, we only consider directed graphs with nonnegative edge weights, node classification is carried out in a semi-supervised setting.  We employ 10 random data splits for all the datasets. Ten folds are generated randomly for each dataset. For the datasets Cornell, Texas, and Wisconsin, we used the settings from~\cite{He2022PyTorchGS}. In all experiments, we used the normalized nonlinear Laplacian and implemented NLSD-GNN with convolution defined as in Equation~\eqref{eq:convequ2}, which means our network may be viewed as the nonlinear Laplacian generalization of ChebNet.

Link prediction typically aims to determine the presence of a connection between two nodes in unsigned and undirected graphs. However, this task extends into more complex scenarios involving signed and/or directed networks. The initial task, known as link sign prediction (SP), predicts the existence of a directed edge from  $v_i$  to $v_j$ and seeks to classify the edge as positive or negative, determining whether  $ (v_i, v_j) \in E^+$  or $ (v_i, v_j) \in E^- $. The next challenge, direction prediction (DP), involves predicting the directionality of the edge, i.e., whether $(v_i, v_j) \in E$  or  $(v_j, v_i) \in E$ , under the assumption that only one direction is valid.
Further complexities are introduced through multi-class prediction tasks. The three-class challenge (3C) considers three possibilities:  $(v_i, v_j) \in E $,  $(v_j, v_i) \in E $, or neither. In the four-class scenario (4C), the predictions expand to include both positive and negative relationships in both directions:  $(v_i, v_j) \in E^+ $,  $(v_i, v_j) \in E^- $,  $(v_j, v_i) \in E^+ $, and  $(v_j, v_i) \in E^-$ . The most complex, the five-class format (5C), adds an additional category where neither edge exists between $v_i$  and  $v_j$.
Performance across these tasks is assessed using classification accuracy. It is noteworthy that while the (SP), (DP), and (3C) tasks require methods capable of discerning either signed or directed aspects, the (4C) and (5C) tasks necessitate methods that can adeptly handle both signs and directions of the edges. 
\vspace{-7pt}

\subsection{Results \& Discussion}

\paragraph{Node Classification.} 
As demonstrated in Table~\ref{tab:result1}, on homophilic datasets, both variants of NLSD-GNN consistently rank among the top three models across the datasets. 
Notably, NLSD-GNN outperforms MagNet and other non-spectral methods, achieving the best results on three out of five datasets and placing second on the Cornell dataset. 
This superior performance underscores the effectiveness of the nonlinear Laplacian in capturing directional information. 
On the Cora-ML dataset, NLSD-GNN places third, slightly behind SAGE. 
In heterophilic datasets presented in Table~\ref{tab:result2}, NLSD-GNN surpasses MagNet and similar spectral GNN methodologies.
However, it falls short against GNNs specifically designed for directed heterophilic environments, such as HoloNet and Dir-GNN detailed in \cite{Koke2023HoloNetsSC, rossi2024edge}. 
Moreover, these targeted heterophilic methods demonstrate minimal gains—and occasionally a slight decline—in performance when applied to homophilic datasets with directional data. 
It is also important to note that the models cited in \cite{Koke2023HoloNetsSC, rossi2024edge} cannot handle signed-directed graphs. Additional results involving various baselines can be found in the Appendix.

\paragraph{Link Prediction on Signed-Directed Graphs.} 

Table~\ref{tab:result3} summarizes the experimental results for link prediction on signed-directed graphs. We observe that NLSD-GNN-Fast outperforms all the baseline methods on almost all the datasets, and MSGNN generally achieves suboptimal results, followed by SSSNET, SNEA and
 SDGNN. In some cases especially For the SP task, it is observed that relying solely on signed attributes, excluding weighted features, yields favorable results.
In summary, developing features that distinctly address the positive and negative subgraphs proves beneficial, and incorporating directional information generally enhances performance. Table~\ref{tab:result3} is instrumental in assessing the performance of the NLSD-GNN model, specifically examining how it handles signs and directions both individually and in combination. This constitutes the core of our ablation study on the NLSD-GNN's capabilities. While simpler tasks such as (SP), (DP), and (3C) evaluate the model's ability to address either signed or directed attributes separately, more intricate tasks like (4C) and (5C) challenge the model to effectively manage both attributes at the same time. We have the following observations: the performance of NLSD-GNN in task (4C) diminishes compared to task (5C), suggesting that effectively quantifying the influence of non-existent links and capturing the structural nuances of signed-directed graphs are crucial for modeling interactions between positive and negative links. Additionally, we observe that integrating directionality into the nonlinear Laplacian matrix generally enhances outcomes in tasks related to directionality (DP, 3C, 4C, 5C). 



%% file: chapters/06_Conclusion.tex
\section{Conclusion}
In this work, we introduced a spectral GNN based on a novel nonlinear signed-directed Laplacian and explored its application in node classification and link prediction tasks that consider both edge sign and directionality. 
The NLSD-GNN not only matches or surpasses the performance of leading GNNs but also operates more efficiently on real-world datasets. 

%% file: chapters/02_RelatedWork.tex
\section{Related Work}
\label{related}
Inspired by GCN architecture, graph neural networks have been extended to directed graphs.
DGCN~\cite{Ma2019} uses a (linear) directed Laplacian associated with random walk, but this restricts the graph to be strongly connected which is not the case in most of the real-world scenarios. 
On the other hand, DiGCN~\cite{tong2020Directed} explores a method using a first-order proximity matrix and two second-order proximity matrices, and incorporates $k$-hop diffusion matrices. 
These matrices enable detailed neighborhood analysis and the construction of multiple Laplacians with a fusion operator for information integration.  
DiGraph~\cite{Tong2020inception} extended these ideas by adapting the directed Laplacian using a personalized PageRank matrix to be suitable for non-strongly connected graphs. 
They also incorporate higher-order receptive fields for enhanced data processing. 
MagNet~\cite{Zhang2021MagNetAN} incorporates directional information in graphs through the use of the magnetic Laplacian, a complex Hermitian matrix. This matrix captures the undirected geometric structure in the magnitude of its entries and encodes directional information in their phase. 
Moreover, \cite{Geisler2023TransformersMD} extends Transformer GNNs for use in directed node/graph classification tasks, while \cite{Maskey2023AFG,maskey2022generalized} has utilized GNNs inspired by ordinary differential equations, extending the concept of oversmoothing to the context of directed graphs.
Recently,~\cite{Koke2023HoloNetsSC} tried to completely remove reliance on the graph Fourier transform and instead take the concept of learnable functions applied to characteristic operators as fundamental.

\begin{table}[!t]
\centering
\caption{Laplacians used in spectral GNNs and their applicability}\label{table:lap_tbl}
\tiny
\begin{tabular}{lccc}
\hline
\textbf{Laplacian} & \textbf{Directed} & \textbf{Signed-Directed} & \textbf{No Extra Param} \\
\hline
Ordinary Laplacian & \(\times\) & \(\times\) & \(\checkmark\) \\
Magnetic Laplacian~\cite{Zhang2021MagNetAN} & \(\checkmark\) & \(\times\) & \(\times\) \\
Signed Laplacian~\cite{hou2003laplacian} & \(\times\) & \(\times\) & \(\checkmark\) \\
Signed magnetic Laplacian~\cite{He2022MSGNNAS} & \(\checkmark\) & \(\checkmark\) & \(\times\) \\
Directed Laplacian~\cite{Yoshida2016NonlinearLF} & \(\checkmark\) & \(\times\) & \(\checkmark\) \\
Signed-directed Laplacian (This work) & \(\checkmark\) & \(\checkmark\) & \(\checkmark\) \\
\hline
\end{tabular}
\end{table}

Recent research primarily explores unsigned graphs, which only include edges of positive signs.
Yet, an expanding focus is on neural networks for signed and directed graphs, aimed at assessing link polarity. 
For instance, SGCN~\cite{derr2018signed} utilizes balance theory~\cite{harary1953notion} for signed-undirected graphs, whereas SiGAT~\cite{huang2019signed}, with a mechanism from~\cite{Velickovic2017GraphAN}, employs a motif-based architecture for signed-directed graph embeddings. 
SDGNN~\cite{huang2021sdgnn} and SNEA~\cite{li2020learning} further developed this area with new efficiencies and objective functions. 
In contrast, SSSNET~\cite{he2022sssnet}, MSGNN~\cite{he2022msgnn}, SLGNN~\cite{Li2023SignedLG} and SigMaNet~\cite{fiorini2023sigmanet} explore semi-supervised clustering and spectral GNNs using signed magnetic Laplacians, respectively. 
Additionally, various GNNs~\cite{singh2022signed, schlichtkrull2018modeling, vashishth2019composition} address multi-relational graphs, which inherently accommodate diverse edge types. 
Signed graphs, particularly those unweighted or with finite weighting scales, offer simplified models of multi-relational graphs where negative edges are as significant as positive, but with fewer parameters needed for network training.

%% file: chapters/x1_appendix1.tex
\section{Proof of Theorem 1 \& 2} \label{app:proof}

Let $(\lambda,v)$ be an eigenpair of the Laplacian $L_G$ of a signed-directed graph $G$.
Suppose $\lambda < 0$.
Then, we have

\begin{equation*}
    0 \leq Q_G(v) = v^\top L_G v = 
    v^\top L_{G_v} v = v^\top (-\lambda) v = -\lambda \|v\|^2 < 0
\end{equation*}
which is a contradiction.
Hence, $L_G$ is positive semidefinite.
A similar argument shows that the normalized Laplacian $\mathcal{L}_G$ is also positive semidefinite.

Now consider an operator $\mathcal{M}: x \mapsto 2x - \mathcal{L}_G(x)$.
To show that all the eigenvalues of $\mathcal{L}_G$ is bounded by two, it suffices to show that $\mathcal{M}$ is positive semidefinite.
This is equivalent to showing that the quadratic form of $\mathcal{M}$ is nonnegative.
For a vertex $v\in V$, let $d_v$ denote the total degree of $v$.
Then, we have

\begin{multline*}
    x^\top \mathcal{M}(x) = 2\|x\|^2 - x^\top D_G^{-1/2}Q_G(x)(D_G^{-1/2}x) \\
    = 2\|x\|^2 - \sum_{e = (u,v) \in E^+}w(e) \max\left\{\frac{x_u}{d_u^{1/2}} - \frac{x_v}{d_v^{1/2}},0\right\}^2 \\
    - \sum_{e = (u,v) \in E^-}w(e) \left(\frac{2x_u^2}{d_u} + \frac{2x_v^2}{d_v} -  \max\left\{\frac{x_u}{d_u^{1/2}} - \frac{x_v}{d_v^{1/2}},0\right\}^2\right) \\
    = \sum_{e = (u,v) \in E^+}w(e) \left(\frac{2x_u^2}{d_u} + \frac{2x_v^2}{d_v} - \max\left\{\frac{x_u}{d_u^{1/2}} - \frac{x_v}{d_v^{1/2}},0\right\}^2\right) \\
    + \sum_{e = (u,v) \in E^-}w(e) \max\left\{\frac{x_u}{d_u^{1/2}} - \frac{x_v}{d_v^{1/2}},0\right\}^2 \\
    \geq 0.
\end{multline*}
Hence, the claim holds. 
Note that the second-to-last expression is the quadratic form of the normalized Laplacian of a signed-directed graph obtained from $G$ by flipping signs of edges.





\section{Further implementation details}

We configured both ChebNet and MagNet and MSGNN with $K = 1$ . Each model trained for up to 1500 epochs, employing early stopping when validation errors ceased decreasing after 500 epochs, applicable to both node classification and link prediction tasks. A dropout layer set at a rate of 0.5 precedes the final linear layer. Model selection was based on peak validation accuracy. Filter counts in the convolutional layers were tuned to \([16, 32, 48]\). Node classification learning rates were tested at \([1e^{-3}, 5e^{-3}, 1e^{-2}]\), and a more conservative rate of \(1e^{-3}\) was applied to link prediction due to the larger sample sizes.  
Adam optimizer and \(L2\) regularization (hyperparameter \(5e^{-4}\)) were utilized to mitigate overfitting. The best test performances were derived by grid-search based on validation accuracies. In link prediction, node features were derived from in-degrees and out-degrees to encode directional information from the adjacency matrix.

For link prediction task on signed-directed graphs, all GNNs are trained for 300 epochs. Loss functions from the original publications are employed for SGCN~\cite{derr2018signed}, SNEA~\cite{li2020learning}, SiGAT~\cite{huang2019signed}, and SDGNN~\cite{huang2021sdgnn}, while cross-entropy loss \(L_{\text{CE}}\) is used for SigMaNet~\cite{fiorini2023sigmanet}, SSSNET~\cite{he2022sssnet}, and MSGNN~\cite{fiorini2023sigmanet}. For each input graph, 20\% of edges are randomly set aside as test edges, and the remaining edges are used for training. Five random splits are generated for each graph. For SigMaNet, SSSNET , and MSGNN, node features are formed into four-dimensional vectors based on the signed in- and out-degrees, excluding test edges. Default settings from \cite{he2024pytorch} are utilized for SGCN , SNEA, SiGAT , and SDGNN. 

Computational experiments were conducted on 1 compute nodes, equipped with 1 Nvidia Quadro RTX 8000 GPU, 380GB RAM, and 32 Intel Xeon E5-2660 v3 CPUs. 

\begin{algorithm*}
\caption{Simple MLP for Graph Feature Reduction}
\begin{algorithmic}[1] 
    \STATE \textbf{Input:} $H$, the initial feature matrix of the graph (dimensions $N \times D$)
    \STATE \textbf{Output:} $x$, the one-dimensional feature vector (dimensions $N \times 1$)
    \STATE Initialize $H_{\text{current}} = H$
    \STATE Set $D_{\text{current}} = D$
    \WHILE{$D_{\text{current}} > 1$} 
        \STATE Halve the dimensionality at each layer: $D_{\text{next}} = \frac{D_{\text{current}}}{2}$
        \FOR{$i = 1$ to $D_{\text{next}}$} 
            \STATE Apply MLP layer: $H_{\text{next}}[:, i] = \tanh(H_{\text{current}} \cdot W[:, i] + b[i])$
        \ENDFOR
        \STATE Update $H_{\text{current}}$ to $H_{\text{next}}$
        \STATE Update $D_{\text{current}}$ to $D_{\text{next}}$
    \ENDWHILE
    \STATE Final transformation: $x = H_{\text{current}} \cdot W_{\text{final}} + b_{\text{final}}$
    \STATE \textbf{Return} $x$
\end{algorithmic}
\end{algorithm*}

\section{Datasets} \label{app:datasets}
\subsection{Node Classification}
As illustrated in Table 5, we employ ten real datasets for node classification. We define a directed edge as follows: if $(u, v) \in E$ but $(v, u) \notin E$, then $(u, v)$ is considered a directed edge. Conversely, if both $(u, v) \in E$ and $(v, u) \in E$, they are treated as undirected edges; undirected edges that are not self-loops are counted twice. For the citation datasets Cora-ML and Citeseer, we randomly select 20 nodes per class for training, 500 nodes for validation, and allocate the remainder for testing, as per. For Chameleon and Squirrel we use the
fixed GEOM-GCN~\cite{Pei2020} splits, for Arxiv-Year we use the splits provided in~\cite{Lim2021LargeSL}, while for Roman-Empire we use the splits from~\cite{Platonov2023ACL}. We report the mean and
standard deviation of the test accuracy, computed over 10 runs in all experiments..


\begin{table*}[!ht]
\centering
\caption{Statistics of the datasets used in Node Classification Task.}
\label{tab:dataset_statistics}
\begin{tabular}{lrrrrrr}
\toprule
{Dataset} & {\# Nodes} & {\# Edges} & {\# Feat.} & {\# C} & {\% Unidirectional Edges} & {Edge Hom.} \\
\midrule
CiteSeer & 4,230 & 5,358 & 602 & 6 & 99.61 & 0.949 \\
Cora-ML & 2,995 & 8,416 & 2,879 & 7 & 96.84 & 0.792 \\
Chameleon & 2,277 & 36,101 & 2,325 & 5 & 85.01 & 0.235 \\
Squirrel & 5,201 & 217,073 & 2,089 & 5 & 90.60 & 0.223 \\
Arxiv-year & 169,343 & 1,166,243 & 128 & 40 & 99.27 & 0.221 \\
Roman-Empire & 22,662 & 44,363 & 300 & 18 & 65.24 & 0.050 \\
Cornell & 183 & 295 & 1,703 & 5 & 86.90 & 0.050 \\
Texas & 183 & 309 & 1,703 & 5 & 76.60 & 0.050 \\
Wisconsin & 251 & 499 & 1,703 & 5 & 77.90 & 0.050 \\
\bottomrule
\end{tabular}
\end{table*}

\begin{table*}[!ht]
\centering
\caption{Summary of various datasets with link statistics for Signed-Directed Link prediction}
\label{tab:datasets}
\begin{tabular}{lrrrr}
\toprule
{Dataset} & {\# Nodes} & {\# Links} & {\% Positive Links} & {\% Negative Links} \\
\midrule
Bitcoin-OTC & 5,872 & 21,431 & 85.29 & 14.71 \\
Bitcoin-Alpha & 3,772 & 14,077 & 90.69 & 9.31 \\
Slashdot & 82,140 & 549,202 & 77.40 & 22.6 \\
Epinions & 131,828 & 841,372 & 85.30 & 14.7 \\
\bottomrule
\end{tabular}
\end{table*}

\subsection{Link prediction on Signed-Directed graphs}
We assess the performance of our NLSD-GNN using four well-known signed-directed graph datasets. The datasets include Bitcoin-Alpha and Bitcoin-OTC, which are networks of Bitcoin traders who assign trust or distrust labels to each other. Another dataset is Slashdot, a network where individuals label one another as friends or foes, derived from interactions on the Slashdot tech news site. The fourth dataset, Epinions, comprises a network from the consumer review site Epinions, where users express trust or distrust towards others. These datasets are summarized in Table 6, highlighting the significant imbalance between the numbers of positive and negative links, where positive links overwhelmingly predominate.
In the results presented in Table 9, Following the setting of the~\cite{Zhang2021MagNetAN} on the directed graphs, we develop our datasets for training, validation, and testing, which consist of vertex pairs from the graph, we implement the following procedures:
\begin{enumerate}
    \item \textbf{Existence Prediction:} For a pair \((u, v)\), the label is 0 if \((u, v) \in E\), and 1 otherwise. Class proportions are 25\% and 75\% for scenarios including undirected and multi-edges, and 50\% each when considering only directed edges.
    \item \textbf{Direction Prediction:} For an ordered node pair \((u, v)\), we assign label 0 if \((u, v) \in E\) and label 1 if \((v, u) \in E\), provided either \((u, v) \in E\) or \((v, u) \in E\). Both edge types maintain a 50\% proportion.
    \item \textbf{Three-Class Link Prediction:} For a pair \((u, v)\), the labels are 0 if \((u, v) \in E\), 1 if \((v, u) \in E\), and 2 if neither condition is met, with respective proportions of 25\%, 25\%, and 50\%.
    \item \textbf{Direction Prediction by Three Classes Training:} This task builds on the third, evaluating only when \((u, v) \in E\) or \((v, u) \in E\).
\end{enumerate}
Ten random folds were generated for all datasets. We allocate 15\% and 5\% of edges for testing and validation across all datasets.
Also, we concatenate the original features, with the in-degree and out-degree as the node features in order to allow the models to
learn structural information from the adjacency matrix directly. The connectivity is maintained by
getting the undirected minimum spanning tree before removing edges for validation and testing. For
the results in the main text, undirected edges and, if they exist, pairs of vertices with multiple edges
between them, may be placed in the training/validation/testing sets.


\begin{table*}[t!]
\centering
\scriptsize
\caption{Node classification accuracy (\%) and its standard deviation for real-world directed heterophilic graphs. The numbers below network names indicate the degree of homophiliy.
Top three models are color-coded  as \textcolor{OliveGreen}{First}, \textcolor{BlueViolet}{Second}, \textcolor{Maroon}{Third}.}
\label{tab:result2}
\begin{tabular}{@{}llccccc@{}}
\toprule
Type & Method & Squirrel & Chameleon & Arxiv-Year & Roman-Empire \\ 
\multicolumn{1}{r}{} & & \textit{0.223} & \textit{0.235} & \textit{0.221} & \textit{0.05} \\ 
\midrule
\multirow{2}{*}{Baseline} & MLP & 28.77 \tiny{$\pm$ 1.56} & 46.21 \tiny{$\pm$ 2.99} & 36.70 \tiny{$\pm$ 0.21} & 64.94 \tiny{$\pm$ 0.62} \\
& GCN & 53.43 \tiny{$\pm$ 2.01} & 64.82 \tiny{$\pm$ 2.24} & 46.02 \tiny{$\pm$ 0.26} & 73.69 \tiny{$\pm$ 0.74} \\
\midrule
\multirow{7}{*}{NON-GNN} & H2GCN & 37.90 \tiny{$\pm$ 2.02} & 59.39 \tiny{$\pm$ 1.98} & 49.09 \tiny{$\pm$ 0.10} & 60.11 \tiny{$\pm$ 0.52} \\
& GPR-GNN & 54.35 \tiny{$\pm$ 0.87} & 62.85 \tiny{$\pm$ 2.90} & 45.07 \tiny{$\pm$ 0.21} & 64.85 \tiny{$\pm$ 0.27} \\
& LINKX & 61.81 \tiny{$\pm$ 1.80} & 68.42 \tiny{$\pm$ 1.38} & 56.00 \tiny{$\pm$ 0.17} & 37.55 \tiny{$\pm$ 0.36} \\
& FSGNN & 74.10 \tiny{$\pm$ 1.89} & 78.27 \tiny{$\pm$ 1.28} & 50.47 \tiny{$\pm$ 0.21} & 79.92 \tiny{$\pm$ 0.56} \\
& ACM-GCN & 67.40 \tiny{$\pm$ 2.21} & 74.76 \tiny{$\pm$ 2.20} & 47.37 \tiny{$\pm$ 0.59} & 69.66 \tiny{$\pm$ 0.62} \\
& GLOGNN & 57.88 \tiny{$\pm$ 1.76} & 71.21 \tiny{$\pm$ 1.84} & 54.79 \tiny{$\pm$ 0.25} & 59.63 \tiny{$\pm$ 0.69} \\
& Grad. Gating & 64.26 \tiny{$\pm$ 2.38} & 71.40 \tiny{$\pm$ 2.38} & 63.30 \tiny{$\pm$ 1.84} & 82.16 \tiny{$\pm$ 0.78} \\
\midrule
\multirow{4}{*}{Directed} & DIGCN & 37.74 \tiny{$\pm$ 1.54} & 52.24 \tiny{$\pm$ 3.65} & OOM & 52.71 \tiny{$\pm$ 0.32} \\
& MagNeT & 39.01 \tiny{$\pm$ 1.93} & 58.22 \tiny{$\pm$ 2.87} & 60.29 \tiny{$\pm$ 0.27} & 88.07 \tiny{$\pm$ 0.27} \\
& DIR-GNN & \textcolor{BlueViolet}{75.31 \tiny{$\pm$ 1.92}} & \textcolor{BlueViolet}{79.71 \tiny{$\pm$ 1.26}} & \textcolor{BlueViolet}{64.08 \tiny{$\pm$ 0.26}} & \textcolor{BlueViolet}{91.23 \tiny{$\pm$ 0.32}} \\
& HoloNet & \textcolor{OliveGreen}{76.71 \tiny{$\pm$1.92}} & \textcolor{OliveGreen}{80.33 \tiny{$\pm$ 1.19}} & \textcolor{OliveGreen}{64.43 \tiny{$\pm$ 0.28}} & \textcolor{OliveGreen}{92.24 \tiny{$\pm$ 0.43}} \\
\midrule
\multicolumn{2}{l}{\textbf{NLSD-GNN-Fast (ours)}} & \textcolor{Maroon}{68.34 \tiny{$\pm$ 1.37}} & \textcolor{Maroon}{75.27 \tiny{$\pm$ 1.42}} & \textcolor{Maroon}{62.13 \tiny{$\pm$ 0.1}} & \textcolor{Maroon}{88.22 \tiny{$\pm$ 0.43}} \\
\bottomrule
\end{tabular}
\end{table*}

\begin{table*}[!htb]
\centering
\small
\caption{ Comparison of link prediction test performance across different variants of NLSD-GNN. The ablation study investigates the effects of incorporating signed and weighted features, where "T" and "F" denote "True" and "False," respectively. For weighted features, "T" indicates the sum of the entries in the adjacency matrix, whereas "T'" represents the sum of their absolute values.}
\label{tab:link_task_data}
\begin{tabular}{llccccc}
\toprule
Data Set & Link Task & (F, F) & (F, T) & (F, T') & (T, F) & (T, T) \\
\midrule
\multirow{5}{*}{BitCoin-Alpha} & SP & 71.6±0.9 & 71.6±1.6 & 70.2±1.1 & 71.5±0.9 & 72.7±0.8 \\
                       & DP & 74.8±1.0 & 71.8±1.6 & 71.6±1.8 & 70.3±0.5 & 70.5±0.5 \\
                       & 3C & 85.8±0.9 & 82.2±0.4 & 84.4±0.6 & 83.6±0.6 & 85.6±1.6 \\
                       & 4C & 58.8±1.1 & 54.2±0.3 & 56.6±1.6 & 58.4±1.4 & 58.7±0.3 \\
                       & 5C & 83.3±0.6 & 81.0±0.5 & 81.9±0.5 & 83.2±0.3 & 82.6±0.3 \\
\midrule
\multirow{5}{*}{BitCoin-OTC}   & SP & 73.7±1.5 & 72.0±0.3 & 73.3±0.8 & 74.1±0.7 & 73.8±0.4 \\
                       & DP & 74.8±0.7 & 74±0.1 & 72.6±1.4 & 75.2±0.4 & 71.8±0.7 \\
                       & 3C & 85.0±0.5 & 84.9±0.8 & 83.3±1.0 & 84.8±0.9 & 84.2±1.7 \\
                       & 4C & 61.4±0.4 & 57.4±0.9 & 55.9±2.1 & 64.5±0.4 & 60.5±0.8 \\
                       & 5C & 82.4±0.8 & 77.0±2.6 & 80.0±0.7 & 81.8±0.4 & 82.5±0.3 \\
\midrule                       
\multirow{5}{*}{Slashdot}      & SP & 93.1±0.0 & 92.0±0.1 & 93.0±0.1 & 92.8±0.1 & 93.2±0.6 \\
                       & DP & 92.1±0.1 & 91.1±0.1 & 93.1±0.1 & 92.0±0.1 & 93.3±0.1 \\
                       & 3C & 84.2±0.2 & 85.3±0.4 & 86.2±0.3 & 85.2±0.2 & 86.8±0.5 \\
                       & 4C & 72.3±1.1 & 72.1±0.1 & 70.7±1.2 & 79.0±0.6 & 79.1±1.3 \\
                       & 5C & 74.8±0.3 & 72.1±0.6 & 72.8±0.3 & 78.5±0.6 & 78.4±0.3 \\
\midrule                       
\multirow{5}{*}{Epinions}      & SP & 85.4±0.1 & 86.6±0.1 & 86.4±0.1 & 84.6±0.2 & 86.1±0.4 \\
                       & DP & 86.1±0.5 & 86.1±0.7 & 86.1±0.4 & 83.3±0.1 & 86.7±0.6 \\
                       & 3C & 83.5±0.2 & 83.3±0.3 & 83.6±0.4 & 82.6±0.3 & 85.2±0.7 \\
                       & 4C & 76.7±0.3 & 77.3±0.3 & 76.2±0.4 & 79.1±1.5 & 79.7±0.1 \\
                       & 5C & 78.5±0.5 & 78.1±0.2 & 77.3±0.2 & 80.6±0.2 & 81.3±0.1 \\
\bottomrule
\end{tabular}
\end{table*}

\begin{table*}[t!]
 \centering
 \scriptsize
 \caption{Link prediction accuracy (\%) for directed graph. Top three models are color-coded as \textcolor{OliveGreen}{First}, \textcolor{BlueViolet}{Second}, \textcolor{Maroon}{Third}.  }\label{tab:result2}
 \begin{tabular}{l@{\quad}llll@{\quad}llll}
    \toprule
    \multirow{2}{*}{Method}
    &\multicolumn{4}{c}{Direction prediction} & \multicolumn{4}{c}{Existence prediction} 
    \\\cmidrule(lr){2-5}\cmidrule(lr){6-9}
     &Cornell & Wisconsin & Cora-ML& CiteSeer & Cornell & Wisconsin & Cora-ML& CiteSeer \\
     \midrule
     GCN &56.2 \tiny{$\pm$ }8.7 &71.0 \tiny{$\pm$ }4.0 &79.8 \tiny{$\pm$ }1.1 &68.9 \tiny{$\pm$ }2.8 &75.1 \tiny{$\pm$ }1.4 &75.1 \tiny{$\pm$ }1.9& 81.6 \tiny{$\pm$ }0.5 &76.9 \tiny{$\pm$ }0.5\\
    ChebNet &71.0 \tiny{$\pm$ }5.5& 67.5 \tiny{$\pm$ }4.5 &72.7 \tiny{$\pm$ }1.5 &68.0 \tiny{$\pm$ }1.6& 80.1 \tiny{$\pm$ }2.3 &82.5 \tiny{$\pm$ }1.9&80.0 \tiny{$\pm$ }0.6 &77.4 \tiny{$\pm$ }0.4\\
    \midrule
    APPNP& 69.5 \tiny{$\pm$ }9.0 &75.1 \tiny{$\pm$ }3.5 &83.7 \tiny{$\pm$ }0.7& 77.9 \tiny{$\pm$ }1.6& 74.9 \tiny{$\pm$ }1.5& 75.7 \tiny{$\pm$ }2.2& \textcolor{BlueViolet}{82.5 \tiny{$\pm$ }0.6}& 78.6 \tiny{$\pm$ }0.7\\
    SAGE& 75.2 \tiny{$\pm$ }11.0& 72.0 \tiny{$\pm$ }3.5& 68.2 \tiny{$\pm$ }0.8& 68.7 \tiny{$\pm$ }1.5 &79.8 \tiny{$\pm$ }2.4& 77.3 \tiny{$\pm$ }2.9& 75.0 \tiny{$\pm$ }0.0 &74.1 \tiny{$\pm$ }1.0\\
    GIN& 69.3 \tiny{$\pm$ }6.0& 74.8 \tiny{$\pm$ }3.7& 83.2 \tiny{$\pm$ }0.9& 76.3 \tiny{$\pm$ }1.4& 74.5 \tiny{$\pm$ }2.1& 76.2 \tiny{$\pm$ }1.9& 82.5 \tiny{$\pm$ }0.7& 77.9 \tiny{$\pm$ }0.7\\
    GAT& 67.9 \tiny{$\pm$ }11.1& 53.2 \tiny{$\pm$ }2.6& 50.0 \tiny{$\pm$ }0.1& 50.6 \tiny{$\pm$ }0.5& 77.9 \tiny{$\pm$ }3.2& 74.6 \tiny{$\pm$ }0.0& 75.0 \tiny{$\pm$ }0.0& 75.0 \tiny{$\pm$ }0.0\\
    \midrule
    DGCN& \textcolor{Maroon}{80.7 \tiny{$\pm$ }6.3}& 74.5 \tiny{$\pm$ }7.2& 79.6 \tiny{$\pm$ }1.5& 78.5 \tiny{$\pm$ }2.3& 80.0 \tiny{$\pm$ }3.9& \textcolor{BlueViolet}{82.8 \tiny{$\pm$ }2.0}& 82.1 \tiny{$\pm$ }0.5& \textcolor{BlueViolet}{81.2 \tiny{$\pm$ }0.4}\\
    DiGraph& 79.3 \tiny{$\pm$ }1.9& 82.3 \tiny{$\pm$ }4.9& 80.8 \tiny{$\pm$ }1.1& 81.0 \tiny{$\pm$ }1.1& \textcolor{OliveGreen}{80.6 \tiny{$\pm$ }2.5}& \textcolor{BlueViolet}{82.8 \tiny{$\pm$ }2.6}& 81.8 \tiny{$\pm$ }0.5 &\textcolor{OliveGreen}{82.2 \tiny{$\pm$ }0.6}\\
    DiGraphIB &79.8 \tiny{$\pm$ }4.8& 82.0 \tiny{$\pm$ }4.9& 83.4 \tiny{$\pm$ }1.1& 82.5 \tiny{$\pm$ }1.3& \textcolor{BlueViolet}{80.5 \tiny{$\pm$ }3.6}& 82.4 \tiny{$\pm$ }2.2& 82.2 \tiny{$\pm$ }0.5& \textcolor{Maroon}{81.0 \tiny{$\pm$ }0.5}\\
    MagNet& \textcolor{Maroon}{80.7 \tiny{$\pm$ }2.7}&  \textcolor{Maroon}{83.6 \tiny{$\pm$ }2.8}&  \textcolor{Maroon}{86.1 \tiny{$\pm$ }0.9}&  \textcolor{Maroon}{85.1 \tiny{$\pm$ }0.8} &\textcolor{OliveGreen}{80.6 \tiny{$\pm$ }3.8}& \textcolor{OliveGreen}{82.9 \tiny{$\pm$ }2.6}& \textcolor{OliveGreen}{82.8 \tiny{$\pm$ }0.7}& 79.9 \tiny{$\pm$ }0.5\\
    \midrule
    \textbf{NLSD-GNN-MLP} & \textcolor{BlueViolet}{88.4 \tiny{$\pm$ }3.2}& \textcolor{BlueViolet}{89.9 \tiny{$\pm$ }4.2}& \textcolor{OliveGreen}{89.8 \tiny{$\pm$ }2.3
    }& \textcolor{OliveGreen}{92.2 \tiny{$\pm$ }3.0}& \textcolor{Maroon}{80.4 \tiny{$\pm$ }3.2}& 81.9 \tiny{$\pm$ }1.1 & \textcolor{OliveGreen}{82.8 \tiny{$\pm$ }3.3}& 80.1 \tiny{$\pm$ }0.7\\\cline{2-9}
    \textbf{NLSD-GNN-Fast} & \textcolor{OliveGreen}{89.1 \tiny{$\pm$ }5.6}& \textcolor{OliveGreen}{90.3 \tiny{$\pm$ }4.0}& \textcolor{BlueViolet}{89.3 \tiny{$\pm$ }0.1}& \textcolor{BlueViolet}{90.1 \tiny{$\pm$ }0.3}& 79.9 \tiny{$\pm$ }1.2& \textcolor{Maroon}{82.5 \tiny{$\pm$ }4.7}& \textcolor{Maroon}{82.4 \tiny{$\pm$ }2.3}& 79.8 \tiny{$\pm$ }1.5\\
\midrule

\end{tabular}
\end{table*}

\section{Ablation Study and discussion}
Table 8 compares different variants of NLSD-GNN on the link prediction tasks. In our analysis, we evaluate the impact of including sign information in input node features and the consideration of edge weights. Typically, the calculation of degrees with signed edge weights yields net degrees, where edge weights are summed directly. This approach allows opposing signs, such as -1 and +1, to neutralize each other. In contrast, the features aggregating absolute values of edge weights are labeled as T' in the table. Each configuration is represented by a tuple: (inclusion of signed features, inclusion of weighted features), where "T" and "F" signify "True" and "False," respectively. Here, "T" indicates the sum of entries in the adjacency matrix, whereas "T'" represents the sum of the absolute values of these entries. Based on the results there is no significant difference between the two options, T and "T'", as weight features when signed features are excluded. Also, the results suggest that the use of unit-magnitude weights may be either advantageous or detrimental, varying by dataset and task. 

Moreover, We achieve the best performance on the direction prediction task on all of the datasets as shown in Table 9. In comparison to the MagNet, NLSD-GNN obtains a $\sim$ 9\% improvement on Cornell and a $\sim$ 7\% improvement on Wisconsin, while on Cora-ML and CiteSeer it improves by $\sim$ \%3 and \%5 respectively. 
This emphasizes the role of the nonlinear Laplacian in our model.  On the existence prediction, while MagNet has a better performance than our model, our model in three out of four datasets, is still in the top three models narrowly behind the second.  

\section{Complexity Analysis} \label{app:complex}

In this section, we evaluate the efficiency of our model against some of the directed baselines. 
Assuming the DirGCN configuration for the DirGNN baseline (with higher complexities possible for other configurations), consider a graph \( G \) with \( N \) nodes and \( M \) edges, input features of dimension  $d$ , and \( F \) hidden dimensions in our architectures. The computational complexities are as follows:

\begin{itemize}
    \item \textbf{DiGCN:} Typically, In its full form, DiGCN has a $O(N^2)$ complexity, which is reduced to a \( O(MFd) \) complexity
in its approximate propagation scheme, which is used in practice.
    \item  \textbf{DirGNN:} Similarly, DirGNN possesses a \( O(MFd) \). In both cases this stems from the sparse-dense matrix multiplications that facilitate the forward function.
    \item \textbf{ MagNet, MSGNN:} Both implement their forward functions through sparse-dense matrix multiplications, resulting in \( O(MFd) \) complexity.
    \item \textbf{ NLSD-GNN:}  NLSD-GNN is similarly implemented via sparse-dense matrix multiplications; resulting in an
\( O(MFd) \) complexity.
\end{itemize}

\textbf{Number of trainable parameters:} 

Compared to MSGNN, our \textbf{NLSD-GNN-MLP}, assuming equivalent network width and depth, introduces a slightly higher number of learnable parameters primarily in the MLP part. 
However, this increase is negligible, especially when employing a shallow MLP configuration in practical scenarios. Furthermore, for the \textbf{NLSD-GNN-FAST} variant, the total count of parameters aligns closely with that of MSGNN, showing no significant difference in parameter complexity between these models. in Algorithm 2, we present a simple pseudocode for how we can use an MLP for feature reduction to obtain a one-dimensional vector for use in our non-linear Laplacian. This demonstrates that enhancements in our model do not necessarily come at the cost of increased computational overhead.

\section{Limitations and Ethical Considerations:}
In this work, we introduced a spectral GNN based on a novel nonlinear signed-directed Laplacian and explored its application in node classification and link prediction tasks that consider both edge sign and directionality. 
The NLSD-GNN not only matches or surpasses the performance of leading GNNs but also operates more efficiently on real-world datasets. 
Future work will investigate additional properties of our proposed Laplacian and extend its application to temporal and dynamic graphs, where node features and edge information may evolve over time. \\

Our method extends spectral graph convolutional networks to signed-directed graphs, though scalability to larger graphs remains a challenge for future development. Additionally, the potential of NLSD-GNN in heterophilic graph contexts has not been explored, indicating further avenues for research. The societal impact of this method aligns with that of other graph neural network algorithms, neither significantly greater nor lesser.